\pdfoutput=1

\documentclass[11pt]{article}

\usepackage[]{ACL2023}

\usepackage{times}
\usepackage{latexsym}

\usepackage[T1]{fontenc}

\usepackage[utf8]{inputenc}

\usepackage{microtype}

\usepackage{inconsolata}

\usepackage[utf8]{inputenc} %
\usepackage[T1]{fontenc}    %
\usepackage{hyperref}       %
\usepackage{url}            %
\usepackage{booktabs}       %
\usepackage{amsfonts}       %
\usepackage{nicefrac}       %
\usepackage{microtype}      %
\usepackage{xcolor}         %
\usepackage{amsmath}
\usepackage{amssymb}
\usepackage{amsthm}

\usepackage{url}
\usepackage{microtype}
\usepackage{graphicx}
\usepackage{booktabs} %
\usepackage{colortbl}
\usepackage{tabulary}
\usepackage{adjustbox}
\usepackage{multirow}
\usepackage{makecell} 
\usepackage{caption}
\usepackage{subcaption}
\usepackage{enumitem}
\usepackage{xcolor}

\usepackage{cleveref}
\usepackage{wrapfig}
\usepackage{stackengine}
\usepackage{mathtools}
\usepackage{natbib}
\usepackage{floatrow}

\usepackage[ruled]{algorithm2e}

\usepackage{xcolor}
\hypersetup{
    colorlinks,
    linkcolor={red!50!black},
    citecolor={blue!50!black},
    urlcolor={blue!80!black}
}

\usepackage{bm, dsfont}

\newcommand{\cB}{\mathcal{B}}

\newcommand{\cD}{\mathcal{D}}

\newcommand{\argmin}{\mathop{\mathrm{argmin}}}

\makeatletter
\newcommand{\ve}{\@ifnextchar\bgroup{\velong}{{\bm{e}}}}
\newcommand{\velong}[1]{{\bm{#1}}}
\makeatother

\def\vtheta{{\bm{\theta}}}

\newcommand{\tf}[1]{\textbf{#1}}

\setenumerate[1]{itemsep=0pt,partopsep=0pt,parsep=\parskip,topsep=0pt}
\setitemize[1]{itemsep=0pt,partopsep=0pt,parsep=\parskip,topsep=0pt}
\setdescription{itemsep=0pt,partopsep=0pt,parsep=\parskip,topsep=0pt}

\title{HiFT: A Hierarchical Full Parameter Fine-Tuning Strategy}

\author{First Author \\
  Affiliation / Address line 1 \\
  Affiliation / Address line 2 \\
  Affiliation / Address line 3 \\
  \texttt{email@domain} \\\And
  Second Author \\
  Affiliation / Address line 1 \\
  Affiliation / Address line 2 \\
  Affiliation / Address line 3 \\
  \texttt{email@domain} \\}
\author{Yongkang Liu$^{1,2,5}$, Yiqun Zhang$^{1}$, Qian Li$^3$, Tong Liu$^{2,4}$, \\
  \textbf{Shi Feng$^{1}$, Daling Wang$^{1}$, Yifei Zhang$^{1}$ and Hinrich Schütze$^{2,5}$} \\
        $^1$Northeastern University, China;
        $^2$CIS, LMU Munich, Germany \\
        $^3$Shandong University, China;
        $^4$Institute of Informatics, LMU Munich, Germany \\
        $^5$Munich Center for Machine Learning (MCML), Germany \\
        \texttt{misonsky@163.com,yiqunzhang@stumail.neu.edu.cn,TongLiu.physics@gmail.com} \\
        \texttt{feiwangyuzhou@sdu.edu.cn,\{fengshi,wangdaling,zhangyifei\}@cse.neu.edu.cn }
}

\begin{document}

\newtheorem{theorem}{Theorem}
\newtheorem{Assumption}[theorem]{Assumption}

\maketitle
\begin{abstract}
Full-parameter fine-tuning (FPFT) has become the go-to choice for adapting language models (LMs) to downstream tasks due to its excellent performance. 
As LMs grow in size, fine-tuning the full parameters of LMs requires a prohibitively large amount of GPU memory.
Existing approaches utilize zeroth-order optimizer 
to conserve GPU memory, which potentially compromises the performance of LMs as non-zero order optimizers tend to converge more readily on most downstream tasks.
We propose a novel, memory-efficient, optimizer-independent, end-to-end hierarchical fine-tuning strategy, HiFT, which only updates a subset of parameters at each training step. HiFT significantly reduces the amount of gradients and optimizer state parameters residing in GPU memory at the same time, thereby reducing GPU memory usage. Our results demonstrate that: (1) HiFT achieves comparable performance with parameter-efficient fine-tuning and standard FPFT. (2) Results on six models show that HiFT reduces the number of trainable parameters by about 89.18\% on average compared to FPFT.  (3) HiFT supports FPFT of 7B models for 24G GPU memory devices under mixed precision without using any memory saving techniques. (4) HiFT supports various optimizers including AdamW, AdaGrad, SGD, etc.
The source code link is \url{https://github.com/misonsky/HiFT}.
\end{abstract}
\section{Introduction}
Full-Parameter Fine-Tuning (FPFT) Language Models (LMs) have been a successful paradigm in various downstream tasks~\cite{NIPS20173f5ee243,LiuGGLEGLZ20}. However, as the size of LMs becomes larger, FPFT LMs require immense memory, which has become an obstacle to conducting research.
One line of research to reduce memory is to use heterogeneous memory~\cite{abs200205645,RajbhandariRRSH21} (e.g., GPU, CPU, and NVMe memory) or distributed techniques (e.g., tensor parallelism~\cite{ShazeerCPTVKHLH18,MegatronLM,ZhangYXDTZC23,KimKYC23,YUAN2}). These strategies require parameter sharing across diverse devices and thus usually introduce a significant communication burden.
Parameter-Efficient Fine-Tuning (PEFT) is another line of strategies for memory reduction, categorized into addition-based, selection-based, and reparametrization-based methods~\cite{lialin2023scaling}. The addition-based methods (e.g., Prefix-Tuning~\cite{li2021prefix}, AttentionFusion~\cite{cao2022attention}) reduce the number of trainable parameters by only updating newly added parameters and freezing the weights of LMs. Although these methods reduce the number of parameters for fine-tuning, 
they expand the number of model parameters and increase the burden on forward propagation. The selection-based methods (e.g, BitFit~\cite{zaken2022bitfit}, LT-SFT~\cite{ansell2022composable}, FAR~\cite{vucetic2022efficient}), on the other hand, fine-tune a subset of model parameters, resulting in a performance gap with FPFT. The reparametrization-based methods (e.g., LoRA~\cite{hu2021lora}, KronA~\cite{edalati2022krona}, S4-model~\cite{chen2023parameter}) leverage low-rank decomposition to minimize the number of trainable parameters. Using low-rank representations inevitably leads to information loss and performance degradation. 
PEFT involves a trade-off between serving efficiency and quality.
According to existing works~\cite{Finetuning2023,FinetuningLLMs2023,ComprehensiveFinetuningLLMs2023}, FPFT still maintains advantages in performance on most benchmarks.

\begin{figure*}[!t]
\centering
\includegraphics[width=0.75\textwidth]{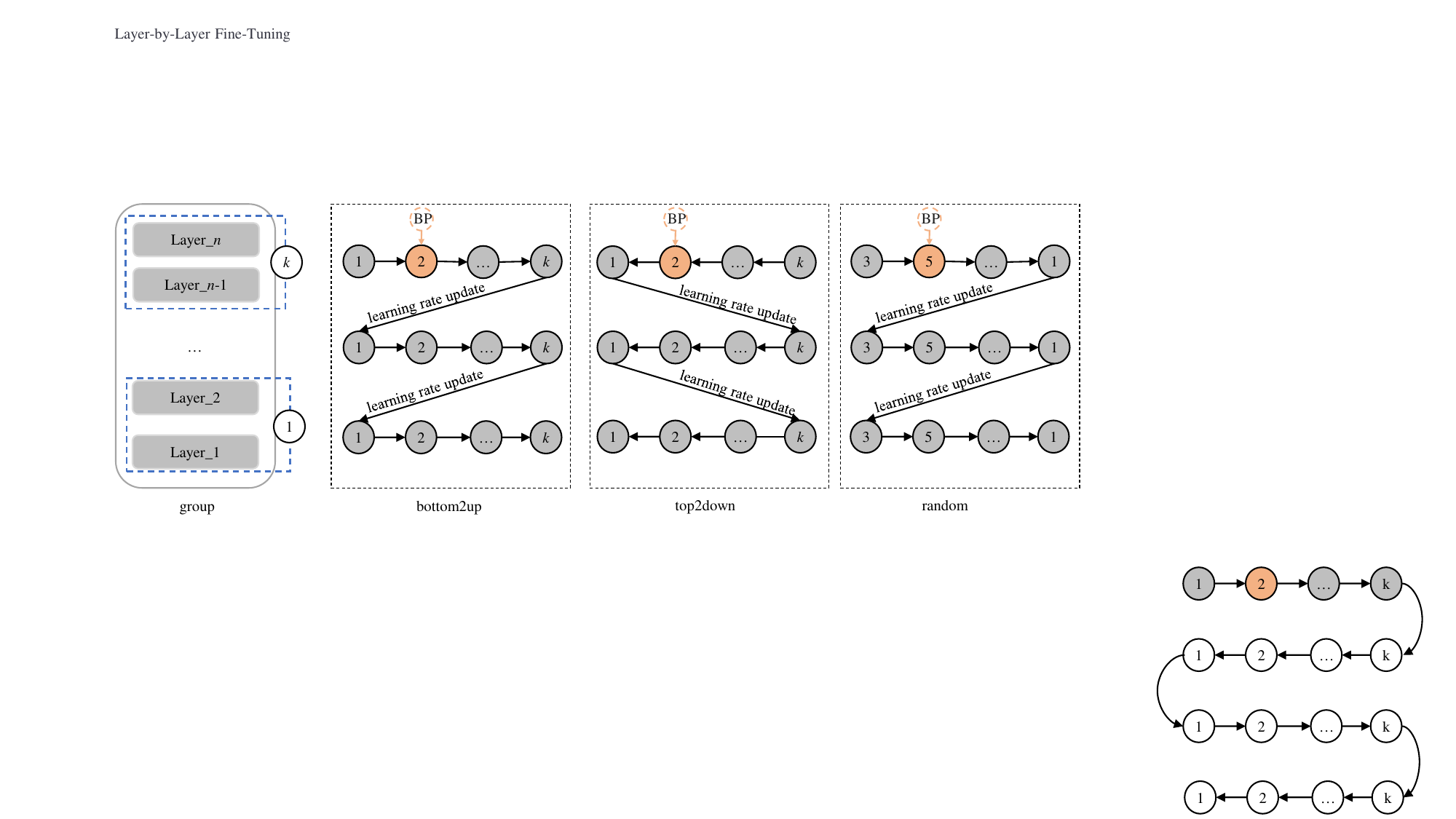}
\vspace{-0.2cm}
\caption{Schematic diagram of our HiFT. \textbf{group} represents the grouping operation of the layers. \textbf{bottom2up}, \textbf{top2down} and \textbf{random} are training strategies. Gray indicates that the corresponding parameters are in the frozen state, and brown indicates that the corresponding parameters are in the activated state. $k$ is the number of groups, $n$ is the number of layers of the given model, and BP denotes parameter update through back propagation.}
\label{fig:framework}
\end{figure*}

Some works have reduced memory usage for FPFT by removing the momentum state of the optimizer. 
LOMO~\cite{LOMO} reduces the memory usage of the optimizer momentum and gradients by integrating gradient calculation and update. Nevertheless, LOMO requires forward propagation twice. In addition, LOMO forces the model to be 16-bit quantized and uses the gradient checkpointing technique~\cite{Chen2016TrainingDN} to reduce memory usage while LOMO has limited memory savings in real-world scenarios.
MeZO~\cite{abs230517333} designs a zeroth-order optimizer to reduce memory usage. However, MeZO is unstable and performs poorly without prompts. These methods make the momentum optimizers unusable, while the momentum optimizers such as AdamW~\cite{Loshchilov2017DecoupledWD} have been proven to be superior in improving performance.

In this paper, we propose a novel memory-efficient \textbf{Hi}erarchical \textbf{F}ine-\textbf{T}uning (\textbf{HiFT}) strategy, adopting the idea of block-by-block training. 
HiFT divides the layers of the model into different groups (a group is a block.). At each training step, HiFT updates the parameters of one group while freezing the others. 
Compared to standard FPFT, HiFT leads to different groups of parameters being updated with different learning rates. This causes the model parameters to be updated in an inconsistent amplitude, which leads to a decrease in model performance. To solve this problem, we adopt to delay the learning rate update, which only updates the learning rate once when all layers of the model are updated. HiFT is also different from layer-wise training~\cite{bengio2006greedy}, where the layer-wise training incrementally adds new layers to a pre-trained shallow model, only updating the newly added parameters at each training stage until all layers are updated. As a result, the layer-wise strategy produces accumulated errors at different training stages due to its pipeline training.

HiFT can significantly reduce the number of trainable parameters per training step. We only keep the momentum and gradients of the parameters that need to be updated on the GPU device due to only a portion of the parameters are updated at each training step. This helps to reduce the GPU memory usage of the optimizer states and gradients. HiFT supports full-parameter fine-tuning of a 7B model on devices with 24G memory.
Our contributions are summarized as follows:
\begin{itemize}[leftmargin=*]
    \item We propose a novel, memory-efficient, optimizer-independent, end-to-end hierarchical fine-tuning strategy HiFT. Different from standard full parameter fine-tuning, HiFT achieves full-parameter fine-tuning in an asynchronous block-by-block manner. 
    \item We show that the order of updates has no impact on model performance during asynchronous block-by-block updates, which provides a basis for block-by-block parallel updates of models in the future.
    \item Experiments show that HiFT achieves the same or even better performance than FPFT and PEFT on instruction fine-tuning, classification, generation, question answering and inference tasks with less GPU memory.
\end{itemize}

\section{Related Work}
\paragraph{\textbf{Full-Parameter Fine-tuning}}
FPFT fine-tunes the pre-trained LMs on specific tasks by updating all parameters~\cite{FullParameterandLoRAbasedFineTuning,lin2024mala,ma2024topro}, which
requires massive computing power as the parameters of LMs increase.
Mixed-precision training enables high-throughput computations by employing half-precision storage for parameters, activations, and gradients~\cite{RajbhandariRRH20,NarayananSCLPKV21}.
Staged training incrementally increases the amount of compute and reuse the compute from prior stages~\cite{ShenWKDPB22}.
These methods increase the parameter consumption when training precision or operators.
LOMO~\cite{LOMO} identifies the memory saving of SGD~\cite{robbins1951stochastic}, fuses the gradient computation and the parameter update in one step.
MeZO~\cite{abs230517333} designs a gradient-free method to update the model. Although it can reduce memory usage, its performance has a big gap than FPFT, especially when there is no prompt.
These methods waste the superiority of momentum optimizers.

\paragraph{\textbf{Parameter-Efficient Fine-tuning}}
PEFT minimizes resource utilization from the perspective of parameters with additon, selection or decomposition methods~\cite{lialin2023scaling}.
The addition-based methods add and update new parameters with the weights of LMs frozen, such as Prefix-Tuning~\cite{li2021prefix}, AttentionFusion~\cite{cao2022attention}, while the added parameters increase the burden on forward propagation. 
The selection-based methods fine-tune a subset of the parameters of LMs, such as BitFit~\cite{zaken2022bitfit}, 
LT-SFT~\cite{ansell2022composable}, FAR~\cite{vucetic2022efficient}, but has a performance gap with FPFT. 
The reparametrization-based methods leverage low-rank decomposition to minimize the number of trainable parameters, such as LoRA~\cite{hu2021lora}, PHM~\cite{karimi2021compacter}, KronA~\cite{edalati2022krona}, S4-model~\cite{chen2023parameter}, while using low-rank representations inevitably leads to information loss and performance degradation.
PEFT involves a trade-off between serving efficiency and quality.
\paragraph{\textbf{Memory-Efficient Fine-tuning}}
MEFT minimizes memory usage with heterogeneous memory (e.g., GPU, CPU and NVMe) or parallel methods (e.g., tensor and pipeline parallelism). 
In a layer-to-layer strategy~\cite{abs200205645}, only the tensors necessary for the computation of a particular layer are transferred to GPU, while the remaining tensors are retained in CPU.
ZeRO-Infinity~\cite{RajbhandariRRSH21} enables the partitioned states and tensors to CPU and NVMe. 
Tensor parallelism accelerates training by parallelizing tensor computations across different GPUs, but requires multiple global communications during each 
propagation~\cite{ShazeerCPTVKHLH18,MegatronLM}. 
Pipeline parallelism accelerates training by breaking the model into segments or layers and processing them sequentially in a pipeline fashion~\cite{ZhangYXDTZC23,KimKYC23,YUAN2}.
These methods transfer massive memory to heterogeneous devices, although temporarily saving memory, still requires a large number of devices.

Different from existing works~\cite{LOMO,abs230517333}, HiFT adopts the idea of block-by-block training to save memory of FPFT, and can be seamlessly integrated with any optimizer.

\section{Approach}
\label{approach}
The training strategy of our \textbf{HiFT} is shown in Figure~\ref{fig:framework}. We first present some necessary notations.
\paragraph{Notation} Given the training dataset $\mathcal{D} = \{(x_i, y_i)\}^{N}_{i=1}$, the goal of the training is to learn a model $M$ with $n$ layers, where $N$ is the number of the training samples, $(x_i, y_i)$ is the labeled data pair. We use $P$ to represent the optimizer, and $\eta_t$ to represent the learning rate schedule. The number of layers in each group is represented by $m$ and the number of groups is represented by $k$. If $m$ is divisible by $n$, then $k = n/m$, otherwise $k = \lfloor n/m \rfloor + 1$. Queue $Q$ is used to store special identifiers that uniquely identify different layers. $S \in$ \{"bottom2up","top2down","random"\} represents the adopted update strategy. 

Consider a pre-trained LM $f_{\theta_{pre}}$ parameterized by $\theta_{pre}$. 
Let $\theta_{fpft}$ and $\theta_{hift}$ denote parameters after full fine-tuning and hierarchical full-parameter fine-tuning after \textit{one training step}, respectively.
Let $\mathcal{L}_{\tau}(\mathcal{D}; \theta)$ be the objective to minimize during fine-tuning, with $\mathcal{D}$ being the input, $\theta$ being updated parameters, and $\tau$ being the task in fine-tuning.
In the process of full fine-tuning, we optimize the model by adjusting its full parameters:

\begin{footnotesize}
\begin{equation}
    \theta_{fpft} = \argmin_{\theta_{pre}} \mathcal{L}_{\tau}(\mathcal{D}; \theta_{pre}), 
\end{equation}
\end{footnotesize}
where the dimension of $\theta_{fpft}$, $|\theta_{fpft}|$, equals the dimension of $\theta_{pre}$, $|\theta_{pre}|$.

In the process of HiFT, only a subset of parameters are updated at one training step. More formally, with optimizing group $i \in \{1, ..., k\}$, we have:
\begin{footnotesize}
\begin{align}
    \theta_{hift}^{(i)} &= \argmin_{\beta_{i} \circ \theta_{hift}^{(i-1)}} \mathcal{L}(\mathcal{D}, \beta_{i} \circ \theta_{hift}^{(i-1)} + (1-\beta_{i}) \circ \theta_{hift}^{(i-1)}) \\
    \theta_{hift}^{(1)} &= \argmin_{\beta_{1} \circ \theta_{pre}} \mathcal{L}(\mathcal{D}, \beta_{1} \circ \theta_{pre} + (1-\beta_{1}) \circ \theta_{pre}), 
\end{align}
\end{footnotesize}
where $\beta_{i}$ denotes a fixed binary mask of parameters, with $\beta_{i} \in \{0, 1\}^{|\theta_{pre}|}$, depending on the training strategy chosen in Figure~\ref{fig:framework}. We simply denote $\theta_{hift}^{(k)}$ as $\theta_{hift}$.

\begin{algorithm}
\footnotesize
\SetKwFunction{select}{SelectParameters}
\SetKwFunction{QueGR}{QueueGetAndRemove}
\SetKwFunction{QueAdd}{QueueAddTail}
\SetKwFunction{UpdateStr}{UpdateStrategy}
\SetKwFunction{UpdateOpt}{UpdateOptimizerParameter}
\SetKwFunction{Forward}{ForwardPropagation}
\SetKwFunction{Backward}{Backpropagation}
\SetKwFunction{MoveGPU}{MoveOptimizerState2GPU}
\SetKwFunction{MoveCPU}{MoveOptimizerState2CPU}
\SetKwFunction{MoveCPU}{MoveOptimizerState2CPU}
\SetKwFunction{IsAllUpdate}{IsAllLayerUpdate}

\textbf{Require}: model $M$ with $n$ layers,
number of layers per group $m$,
batch size $B$,
step budget $T$,
optimizer $P$,
parameter queue $Q$,
update strategy $S$,
learning rate schedule $\eta_t$ \\
\textbf{Initialize}:
Initialize queue $Q$ by layer identifier \\
\UpdateStr{$Q,S$} \\
\For{$t=1,...,T$} {
  a). Freeze all parameters of $M$; \\
  b). Sample batch $\cB\subset \cD$ with random seed $s$ \\ 
  \textit{Select key features of layers to be updated} \\
  c). $E \gets$ \QueGR{$Q,m$}  \\
  \textit{Removed elements added to tail of queue} \\
  d). \QueAdd{$Q,E$}   \\
  e). $\vtheta_s \gets$ \select{$M, E$} \\
  f). Set requires$\_$grad = True of parameters $\vtheta_s$ \\
  g). \UpdateOpt{$P,\vtheta_s$}\\
  h). \Forward{$M,\cB$} \\
  \textit{Preserve optimizer state of $\vtheta_s$ within the GPU} \\
  i). \MoveGPU{$P,\vtheta_s$}  \\
  g). \Backward{$P,\vtheta_s,M$} \& Clear gradients \\
  \textit{Keep optimizer state within the CPU} \\
  k). \MoveCPU{$P,\vtheta_s$}  \\ 
  \If{\IsAllUpdate{$t,n,m$}}{
   Update learning rate $\eta_t$
  }
  \Else{
  Keep the learning rate $\eta_t$ constant
  }
  }
  \caption{HiFT Training Algorithm} 
  \label{al:hift}
\end{algorithm}

\subsection{Hierarchical Training}
FPFT has been proven to achieve the-state-of-art performance in most downstream tasks~\citep{Finetuning2023,FinetuningLLMs2023,ComprehensiveFinetuningLLMs2023}.
Standard FPFT updates all parameters of $M$ at each training step, which requires a large amount of GPU memory to store forward and backward propagation parameters at the same time. Different from standard FPFT, HiFT only updates a part of the model parameters and freezes the remaining parameters at each training step, and achieves fine-tuning of all parameters through block-by-block updates. During the BP process, only the parameters that need to be updated will be stored in the GPU memory, which greatly reduces the GPU memory requirements for FPFT.

\begin{table*}[ht]
\begin{adjustbox}{max width=0.75\textwidth, center}
\includegraphics[width=\textwidth]{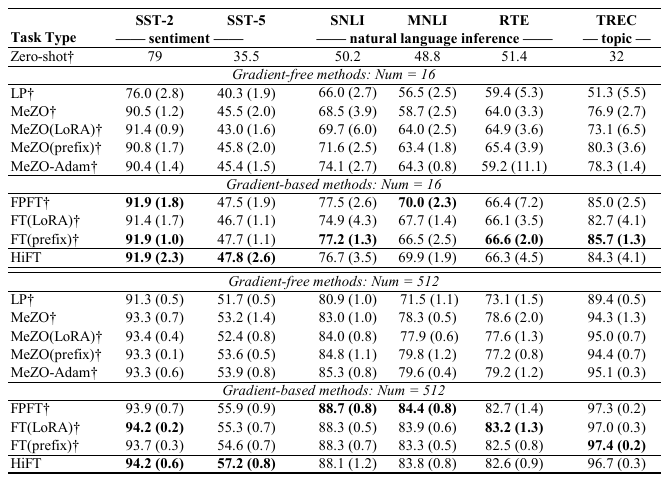}
\end{adjustbox}
\caption{Performance of RoBERTa$_{\text{large}}$ based on prompt fine-tuning. LP: Linear probing; MeZO, MeZO(LoRA) and and MeZO(prefix): memory-efficient ZO-SGD with full-parameter tuning, LoRA, and prefix-tuning respectively; FPFT: fine-tuning with AdamW.  All reported numbers are averaged accuracy (standard deviation). \textit{Num} denotes the number of training examples per class. The parameter $m$ of HiFT is set to 1. \textbf{†} means the result comes from ~\citet{abs230517333}}
\label{tab:r-prompt}
\end{table*}

\begin{table*}[t]
\begin{adjustbox}{max width=0.72\textwidth, center}
\includegraphics[width=\textwidth]{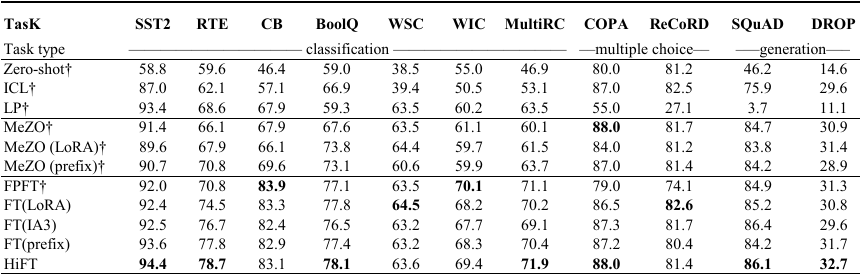}
\end{adjustbox}
\caption{Experiments on OPT-13B (with 1000 examples). ICL: in-context learning; LP: linear probing; FPFT: full fine-tuning; Prefix: prefix-tuning. All experiments use prompts in Appendix~\ref{sec:prompt}. \textbf{†} means the result comes from~\citet{abs230517333}}
\label{tab:opt13-prompt}
\end{table*}

As shown in Figure~\ref{fig:framework}, we divide the model into $k$ groups and update only one group of parameters in each step. All groups are iterated in sequence until convergence. We provide three update strategies: \textbf{bottom2up} (B2U), \textbf{top2down} (T2D) and \textbf{random} (RAN). 
Different strategies only represent different orders of updates, e.g., bottom2up represents the update from the bottom to top. Note that random strategy only shuffles the grouping order before training, and maintains this order in the training process, which avoids the instability caused by constant changes in the update order. Here, the embedding layer is regarded as the bottom layer, and the head layer used for classification or generation is the top layer.

\begin{figure*}[!t]
\centering
\includegraphics[width=0.70\textwidth]{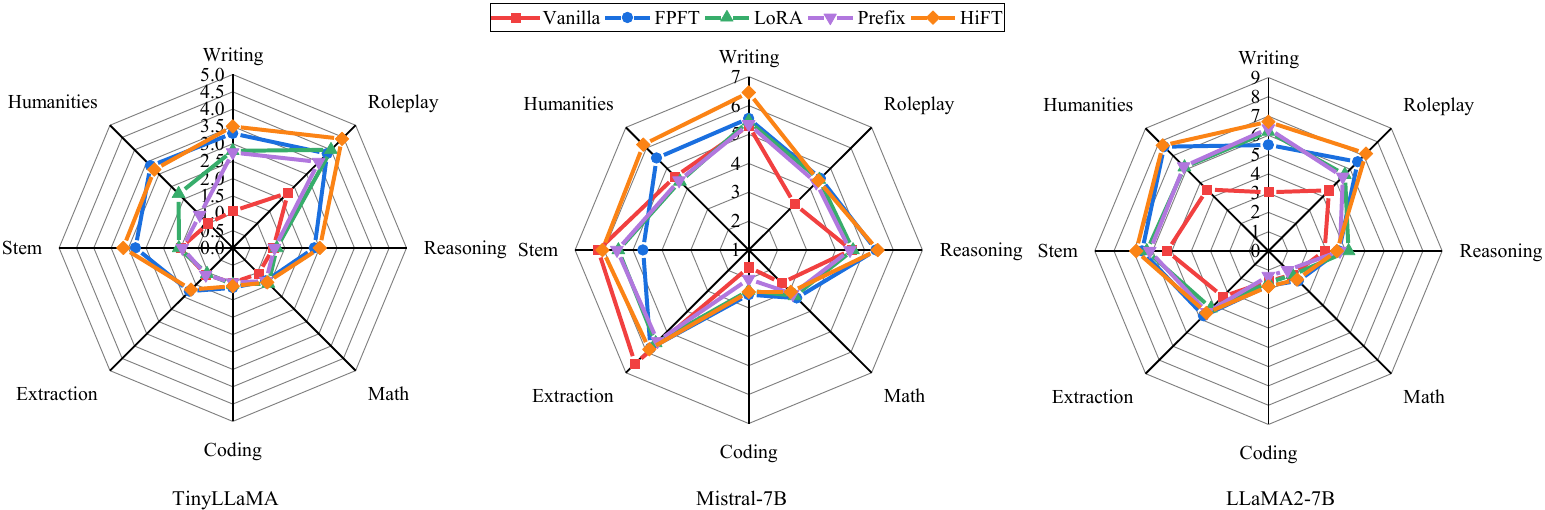}
\caption{Category-wise scores of different fine-tuning methods on MT-bench. The detailed results are shown in Table~\ref{tab:r-mt} (Appendix~\ref{sec:results}).}
\label{fig:radar}
\end{figure*}

\begin{table}[ht]
\begin{adjustbox}{max width=0.85\textwidth, center}
\includegraphics[width=\textwidth]{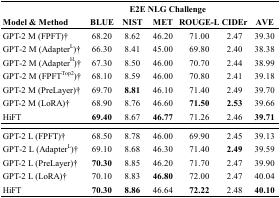}
\end{adjustbox}
\caption{GPT-2 medium (M) and large (L) with different fine-tuning methods on the E2E NLG Challenge. \textbf{†} indicates numbers published in prior works~\cite{gao2024parameter,hu2021lora}.}
\label{tab:gpt2-prompt}
\end{table}

The detailed training process is shown in Algorithm~\ref{al:hift}. The first step is to determine the update strategy.
During training, we freeze all parameters. The layers to be updated, denoted by $E$, are selected from the queue $Q$ based on the parameter $m$. The selected layer $E$ is removed from head of the queue $Q$ and added to the tail of $Q$ to wait for the next update. We select the parameter $\vtheta_s$ that needs to be updated from $M$ based on $E$, set the parameter $\vtheta_s$ to a computable gradient state and set the update parameter group of optimizer $P$ to $\vtheta_s$.
Before parameter updates, the states parameters (e.g., the gradient first moment estimation and second moment estimation of AdamW) of optimizer $P$ related to $\vtheta_s$ could be moved to GPU devices. After the completion of weight updates, the corresponding gradients are cleaned up and optimizer state parameters are moved to CPU. To update the learning rate $\eta_t$, we employ a delayed update strategy. Specifically, we adjust the learning rate once after updating all layers, which helps alleviate the instability issue arising from excessively updates in some layers, especially when fine-tuning deep models. By employing the successive update strategy, the number of parameters residing in GPU simultaneously reduces, thus lowering the GPU memory requirements of fine-tuned models.

Note that we provide a theoretical generalization bound for HiFT (Appendix~\ref{sec:hift-bound}) and a theoretical memory analysis (Appendix~\ref{sec:memory}).

\begin{table}[ht]
\begin{adjustbox}{max width=0.75\textwidth, center}
\includegraphics[width=\textwidth]{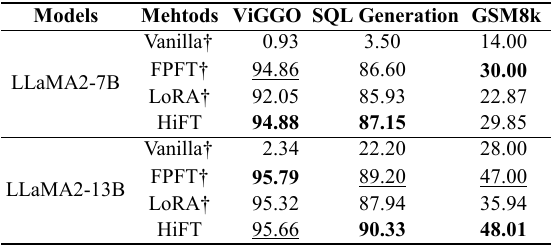}
\end{adjustbox}
\caption{Performance comparison of different fine-tuning methods for LLaMA-7B and 13B. The best result is in bold and the second best result is underlined.}
\label{tab:llama70}
\end{table}

\begin{table*}[ht]
\begin{adjustbox}{max width=0.85\textwidth, center}
\includegraphics[width=\textwidth]{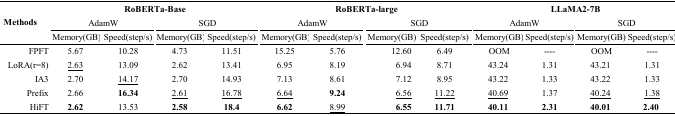}
\end{adjustbox}
\caption{Memory and speed comparison of different fine-tuning methods with mixed precision. The batch size and sequence length are set to 8 and 512. Dataset used by RoBERTa$_{\text{base}}$ and RoBERTa$_{\text{large}}$ is CoLA, and that used by LLaMA2-7B is E2E. All tests were performed on A100 with 80G memory.}
\label{tab:r-ms}
\end{table*}

\section{Experiments}
Please refer to Appendix for baselines(\ref{sec:base}), datasets (\ref{sec:data}) and implementation details (\ref{sec:setting}).

\subsection{Results}
\paragraph{Prompt results} Table~\ref{tab:r-prompt} reports the prompt-based fine-tuning results of the RoBERTa$_\text{large}$. HiFT uses the same prompt template (see Appendix~\ref{sec:prompt}) as MeZO.
We clearly observe that HiFT has an absolute performance advantage compared to gradient-free methods. 
Although gradient-free methods can reduce the memory usage of fine-tuning, there is still a huge gap in performance compared to gradient-based methods. Reducing memory usage at the expense of performance is not an ideal solution. Compared with standard FPFT and PEFT methods, HiFT still achieves competitive results. Table~\ref{tab:opt13-prompt} reports the performance comparison of OPT-13B using different fine-tuning methods on different tasks. We observe that among the 11 tasks, HiFT enjoys performance advantages in 7 tasks. This fully demonstrates the universal effectiveness of HiFT fine-tuning method.

\paragraph{Instruction Fine-tuning} Figure~\ref{fig:radar} and Table~\ref{tab:r-mt} (Appendix~\ref{sec:results}) report the results of instruction fine-tuning for TinyLLaMA, Mistral-7B, and LLaMA2-7B on MT-bench~\cite{zheng2024judging}. We fine-tune these models on Alpaca GPT-4 dataset~\cite{taori2023stanford}. Compared with standard FPFT and PEFT fine-tuning, HiFT has performance advantages in 5 of 8 dimensions on TinyLlaMa, 4 of 8 dimensions on Mistral-7B, and 5 of 8 dimensions on LLaMA2-7B.  In terms of overall performance, HiFT achieves the best results among the three models compared to other fine-tuning methods.

\paragraph{No prompt results} 
Figure~\ref{fig:roberta_result} (Appendix~\ref{sec:results}) shows the performance of RoBERTa$_\text{base}$ and RoBERTa$_\text{large}$ using different fine-tuning strategies on eight tasks. The HiFT performances of RoBERTa$_\text{base}$ have competitive advantages with standard FPFT on datasets such as SST-2, MNLI, QNLI and QQP,
and HiFT has achieved a weak performance advantage on the MRPC dataset. We observe that HiFT has certain performance advantages on most datasets compared to most PEFT methods such as BitFit, Prefix and Adapter. We get similar conclusions on RoBERTa$_\text{large}$. The number of layers of model RoBERTa$_\text{large}$ is about twice that of RoBERTa$_\text{base}$, which reflects to a certain extent that HiFT is not affected by the depth of the model. Table~\ref{tab:gpt2-prompt} reports the results of GPT-2 including medium and large on the E2E dataset. Compared with standard FPFT and PEFT methods, HiFT achieves competitive results on GPT-2 medium and large.
To verify the generalizability of HiFT, we conduct experiments on more complex tasks such as ViGGO~\cite{juraska2019viggo}, SQL generation~\cite{b-mc2_2023_sql-create-context}, and GSM8K~\cite{cobbe2021gsm8k}. Table~\ref{tab:llama70} reports the performance comparison of different fine-tuning methods on these benchmarks. We can observe that HiFT significantly outperforms standard FPFT and LoRA on these three benchmarks. This fully demonstrates the universal effectiveness of HiFT. Another phenomenon is that the performance of LoRA is significantly inferior to standard FPFT and HiFT. To a certain extent, this demonstrates that full parameter fine-tuning is more effective in capturing data characteristics for complex tasks and offers better performance advantages compared to LoRA.
u
\subsection{Memory Efficiency}
To evaluate the effectiveness of HiFT in reducing memory, we compare HiFT with most PEFT methods in terms of memory and speed. Table~\ref{tab:r-ms} reports the memory and speed comparison of different fine-tuning methods on RoBERTa$_\text{base}$, RoBERTa$_\text{large}$ and LLaMA2-7B models. We can observe that HiFT has an absolute advantage in GPU memory usage. HiFT reduces memory usage from three aspects: \textbf{gradients}, \textbf{optimizer states}, and \textbf{residual states}. Since HiFT only updates a small number of parameters in each step, this directly reduces the amount of trainable parameters in each training step, and the corresponding gradient parameters and optimizer state parameters also be reduced in the same proportion. When only some layer parameters are updated in each step, the amount of parameters tracking gradients in the calculation graph is reduced, including the amount of parameters in the activations, so HiFT also reduces the amount of parameters in residual states. This is why HiFT is memory efficient. These PEFT methods introduce new parameters as trainable parameters while freezing the weights of the original LLMs, which reduces the usage of GPU memory by reducing the trainable parameters. Introducing new parameters results in larger memory requirements for the forward computation of fine-tuning. Besides, reducing the number of trainable parameters will reduce the representation ability of models and make them unable to fit complex tasks well. 

We compare LOMO~\cite{LOMO} and MeZO~\cite{abs230517333} based on LLaMA2-7B. Following the settings in Table~\ref{tab:r-ms}, LOMO reports running out of memory on an A100 with 80GB. The memory used by MeZO is about 30GB. MeZO has a memory usage advantage over HiFT due to it being a gradient-free method.
Nevertheless, HiFT significantly outperforms MeZO in terms of performance. Among gradient-based methods, HiFT has advantages in memory.

To evaluate the universality of HiFT in reducing memory, we conduct extensive experiments on different optimizers (i.e., AdamW, SGDM, SGD, Adafactor and Adagrad) based on multiple LMs including RoBERTa$_\text{base}$, RoBERTa$_\text{large}$, GPT-2$_{\text{large}}$, GPT-Neo (2.7B) and LLaMA-2 (7B). Table~\ref{tab:m-base} to Table~\ref{tab:m-llama} (Appendix~\ref{sec:results}) reports the memory usage of the parameters, gradients, optimizer states and residual states under FPFT and HiFT. When using mixed precision, HiFT can save about 44.82\%-53.69\% of memory on RoBERTa$_\text{base}$, about 48.04\%-56.60\% of memory on RoBERTa$_\text{large}$, about 48.20\%-54.27\% of memory on GPT-2$_\text{large}$, about 28.99\%-50.69\% of memory on GPT-Neo and about 65.31\%-76.65\% of memory on LLaMA compared with FPFT.
\begin{figure}[t]
\centering
\includegraphics[width=0.80\textwidth]{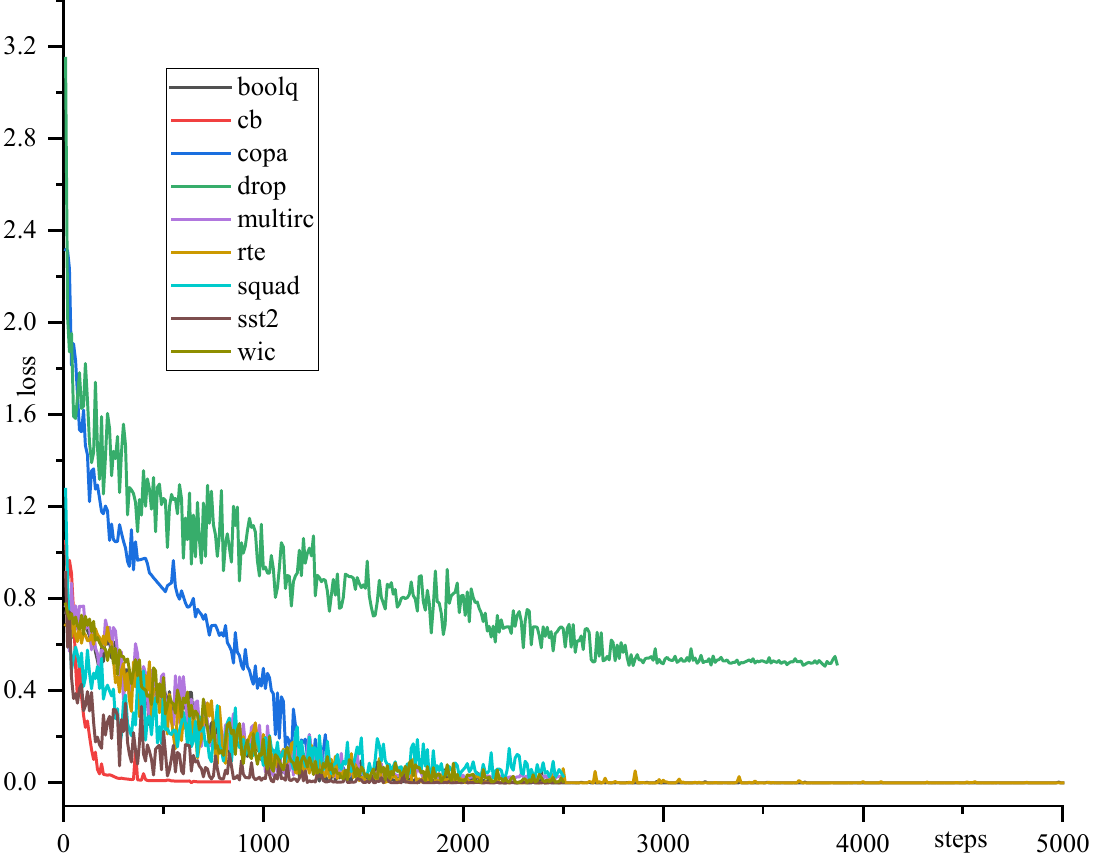}
\caption{Loss curves of OPT-13B on different datasets. The parameter $m$ of HiFT is set to 1.}
\label{fig:loss}
\end{figure}

\subsection{Wallclock Time Efficiency}
In this section, we measure the wallclock time efficiency of HiFT compared to standard FPFT and PEFT methods, with respect to different model sizes. We conduct our experiments on A100 with 80GB GPU memory. Table~\ref{tab:r-ms} reports the wallclock time results for different fine-tuning methods using different optimizers. We can observe that as the number of model parameters increases, the wallclock speed of HiFT gradually gains an advantage. When using the AdamW optimizer, although HiFT is slower than prefix on the RoBERTa$_\text{base}$ model, it is nearly as fast as the prefix method on RoBERTa$_\text{large}$ and faster than PEFT methods on the LLaMA2-7B model. Specifically, on LLaMA2-7B model, HiFT is 1.76× that of LoRA, 1.73× that of IA3, and 1.68× that of prefix. When using the SGD optimizer, HiFT outperforms PEFT and the standard FPFT approach across all models. For LLaMA2-7B model, HiFT is 1.83× that of LoRA, 1.80× that of IA3, and 1.74× that of prefix.

When using the AdamW optimizer, each step of HiFT has a communication cost between the CPU and GPU. The peak communication parameters are shown as the \#Sta values in Table~\ref{tab:m-base} to Table~\ref{tab:m-llama}. The communication cost has limited impact on the speed of HiFT. There are several main reasons: i) The number of communication parameters is small even zero. HiFT is an optimizer-independent method that supports various optimizers. When using SGD, the peak communication parameter is zero. 
When using Adafactor, the peak communication parameter is 0.19MB for RoBERTa$_\text{base}$, 0.21MB for RoBERTa$_\text{large}$, and 0.33MB for LLaMA2-7B. ii) when the required amount of computation reaches the bottleneck of the device, the number of parameters processed per second by the device will no longer increase. Even if the GPU memory is large enough to load parameters, the training speed will not be greatly improved because the computing capability of the device per second is limited. iii) HiFT updates only a subset of parameters at each step, reducing the number of trainable parameters and cutting off gradient propagation to shallow layers. This significantly decreases the computation needed for fine-tuning, thereby increasing the speed. This is why HiFT still has a speed advantage over LLaMA2-7B even with the AdamW optimizer.
\subsection{Stability of Training}
In order to explore the stability of HiFT training, we report the loss curves of OPT-13B on different datasets. As shown in Figure~\ref{fig:loss}, we can observe that during the training process, the loss curve fluctuates within a reasonable range and converges steadily on different datasets. This fully demonstrates that HiFT strategy does not affect the convergence of models. HiFT adopts a delayed learning rate update strategy, which ensures that the update amplitude of parameters in different blocks is consistent and avoids oscillation during the update process.
\subsection{Trainable Parameter}
Figure~\ref{fig:percentage} (e) reports the changes in the amount of peak fine-tuning parameters under HiFT at different model sizes. We observe that as the number of model parameters increases, the proportion of peak trainable parameters gradually decreases. When fine-tuning the 13B model, the peak amount of fine-tunable parameters is only 2.44\% of the original model parameter amount.

Figure~\ref{fig:percentage} shows the percentage of memory used by the parameters of each part when fine-tuning LLaMA2-7B under FPFT and HiFT with the AdamW optimizer. Under FPFT, the optimizer states occupy the most memory. When fine-tuning 32-bit precision (Figure~\ref{fig:percentage} (a)), the memory occupied by residual states is second only to the optimizer state. When mixed precision fine-tuning (Figure~\ref{fig:percentage} (c)), the memory used by model parameters exceeds the memory used by residual states is secondary to the optimizer states. 
The main reason is that in mixed precision training, both 32-bit and half-precision parameters exist at the same time. Therefore, model parameters occupy more memory in mixed precision. HiFT significantly reduces the memory usage of gradients and optimizer states. Therefore, when using HiFT for full-parameter fine-tuning, the main memory-consuming parts are model parameters and residual states.

\begin{figure}[t]
\centering
\includegraphics[width=0.98\textwidth]{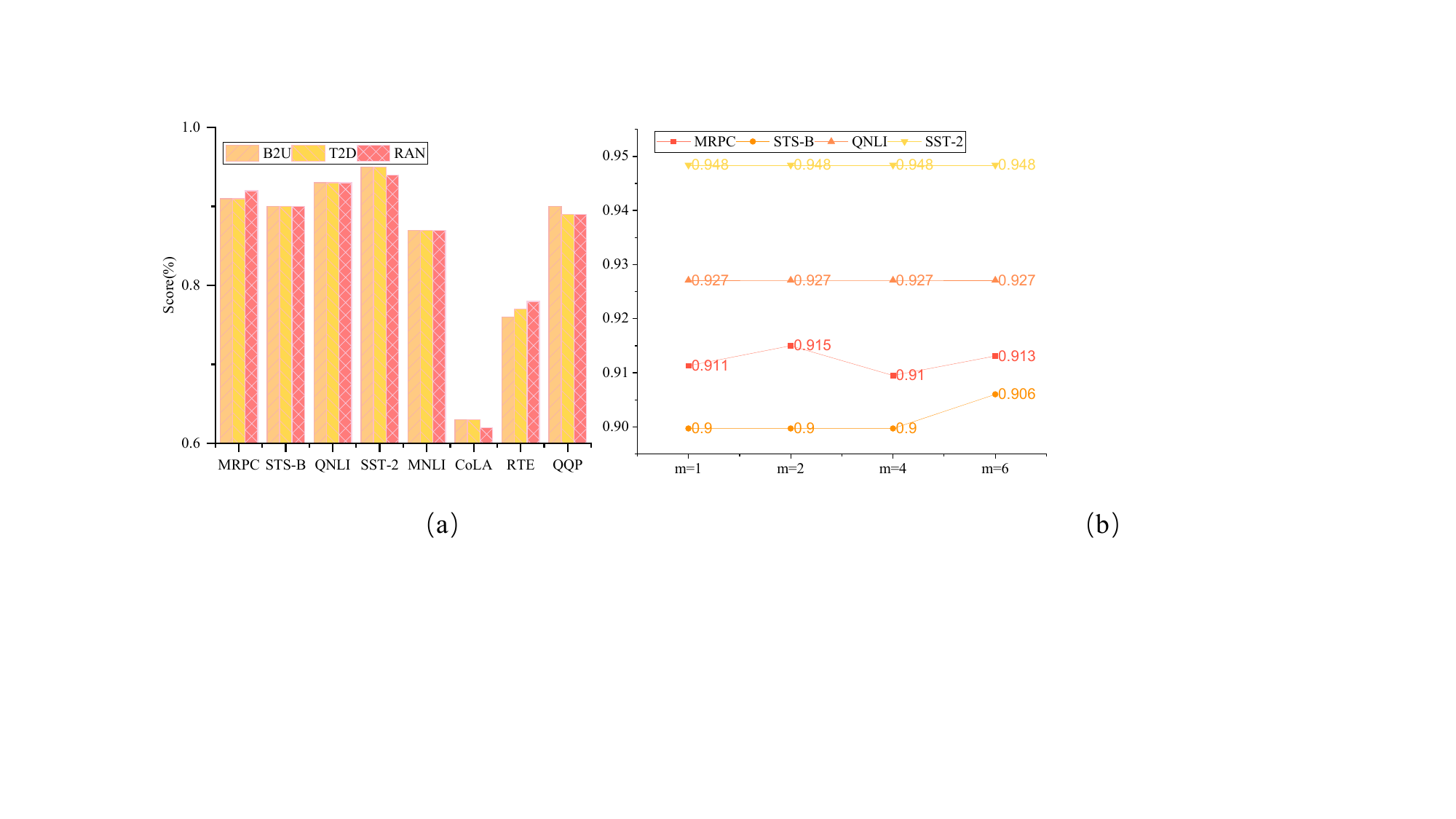}
\caption{The left shows the performance of HiFT of RoBERTa$_\text{base}$ under \textbf{B2U}, \textbf{T2D} and \textbf{RAN} strategies, respectively.
The right shows the performance of HiFT of RoBERTa$_\text{base}$ under different grouping settings, where $m$ is the number of layers in each group.
}
\label{fig:roberta_stra}
\end{figure}

\subsection{Impact of Strategy}
The left plot of Figure~\ref{fig:roberta_stra} reports the performance of RoBERTa$_\text{base}$ using B2U, T2D and RAN strategies.
We observe that the order of updates has almost no effect on the performance of the model. It is an interesting phenomenon that the model still achieves competitive results even when updated in a random order. Changing the update order does not affect the position of the corresponding layer in the model, which is the reason why the performance is not affected. We believe that this phenomenon provides support for hierarchical parallel fine-tuning of large-scale models in the future.
\subsection{Impact of Grouping}
The right plot of Figure~\ref{fig:roberta_stra} reports the impact of different grouping settings on model performance. Although different grouping settings can cause fluctuations in model performance, the overall impact is negligible. We use the learning rate delayed update strategy, which updates the learning rate only after all layers are updated once. This strategy ensures that the learning rate used to update the parameters of each layer is the same in each training step, which helps to prevent the model performance from decreasing due to the update of some parameters being too fast in the hierarchical update process.

\section*{Conclusion}
We propose an end-to-end hierarchical full-parameter fine-tuning strategy, HiFT, which groups the model parameters and updates a single group of parameters per training step. The number of trainable parameters per training step greatly reduce, which lowers the GPU memory usage of the corresponding gradients, optimizer state parameters, and activations. HiFT lowers the barrier of full-parameter fine-tuning of language models and supports full-parameter fine-tuning of a 7B model on a 24G memory device.


\section*{Limitations}
Although HiFT achieves the performance of standard full-parameter fine-tuning at a lower GPU memory cost, there are still some shortcomings. HiFT divides the model by layers, and the maximum division limit is the number of layers of the model. Due to the limitation of the number of layers, HiFT cannot break through the number of model layers for finer-grained division.
When the model width is large, it limits HiFT's capabilities. 
On the other hand, after dividing the model, the number of parameters in each group is different, and the GPU memory usage fluctuates during the fine-tuning process. The peak memory occupied by the fine-tuned model is the decisive factor that determines whether the model is able to be fine-tuned on a certain device. This fluctuation in memory usage during fine-tuning prevents us from fully utilizing resources.

\bibliography{anthology,custom}
\bibliographystyle{acl_natbib}

\clearpage
\appendix

\section{Generalization Bound for HiFT}
\label{sec:hift-bound}
In this section, we establish the generalization bound for HiFT, first building upon a quantization assumption as in \citet{PanigrahiSZA23}. It is important to note that quantization is a common practical consideration; for instance, in our work, we implement a 32-bit quantization precision. 

\begin{Assumption}
\label{assumption:1} 
\emph{(Quantization bound)}
Given model parameters \(\theta\), we denote \(\bar{q}(\theta)\) to be the parameter that quantizes every parameter into the \(q\) given values. Then there exist \(\varepsilon > 0\) s.t. for any sample \(x_{i}\) with label \(y_{i}\) at any training step, we have
\begin{footnotesize}
\begin{equation}
\left| \mathcal{L}((x_{i}, y_{i}); \bar{q}(\theta)) - \mathcal{L}((x_{i}, y_{i}); \theta) \right| \leq \varepsilon.
\end{equation}
\end{footnotesize}
\end{Assumption}

\begin{Assumption}
\label{assumption:2} 
\emph{(Uniform convergence generalization bound for subset parameter fine-tuning) }
Following~\citet{PanigrahiSZA23}, we deviate from the classical uniform convergence generalization bound~\citep{Nagarajan_Uniform_2019} to get a tighter uniform convergence generalization bound for HiFT:

\begin{footnotesize}
\begin{equation}
\begin{split}
    \mathcal{L}_{test}(\theta_{hift}^{(i)}) & - \mathcal{L}_{train}(\theta_{hift}^{(i)}) 
    \\
    & \leq \sup_{\Tilde{\theta}_{hift}^{(i)} \in \Theta} |\mathcal{L}_{test}(\Tilde{\theta}_{hift}^{(i)}) - \mathcal{L}_{train}(\Tilde{\theta}_{hift}^{(i)})|,
\end{split}
\end{equation}
\end{footnotesize}

where $\Theta$ denotes the subset of parameter space, $\theta_{hift}^{(i)}$ being the parameter after i-th optimizing step at one training step. 

\end{Assumption}

\begin{theorem}
\label{theorem:3} 
\emph{(HiFT generalization bound) }
Under Assumption~\ref{assumption:1} and \ref{assumption:2}, we have the following generalization bound for HiFT:  

\begin{footnotesize}
\begin{equation}
\begin{split}
    \mathcal{L}_{test} & (\theta_{hift}^{(k)}) - \mathcal{L}_{test}(\theta^{*}) \\
     & \leq 4k \epsilon + 2 \sum_{i=1}^{k} \sup_{\Tilde{\theta}^{(i)}} | \mathcal{L}_{test}(\bar{q}(\Tilde{\theta}^{(i)})) - \mathcal{L}_{train}(\bar{q}(\Tilde{\theta}^{(i)})) | \\
     & + \mathcal{L}_{test}(\theta^{(k) *}) - \mathcal{L}_{test}(\theta^{*}),
\end{split}
\end{equation}
\end{footnotesize}
where $\theta^{*}$ denotes the parameter with the best test performance, $\Tilde{\theta}^{(i)}$ is in the space of $\beta_{i} \circ \theta_{pre}$ and $\theta^{(i)*}$ denotes the parameter with the best test performance when only changing the subset parameter $\beta_{i} \circ \theta_{pre}$.  
With probability at least $1 - \delta$, the second term $2 \sum_{i=1}^{k} \sup_{\Tilde{\theta}^{(i)}} | \mathcal{L}_{test}(\bar{q}(\Tilde{\theta}^{(i)})) - \mathcal{L}_{train}(\bar{q}(\Tilde{\theta}^{(i)})) |$ can be further bounded: 

\begin{footnotesize}
\begin{equation}
\begin{split}
2 \sum_{i=1}^{k} \sup_{\Tilde{\theta}^{(i)}} & | \mathcal{L}_{test}(\bar{q}(\Tilde{\theta}^{(i)})) - \mathcal{L}_{train}(\bar{q}(\Tilde{\theta}^{(i)})) | 
\\
& \leq 2\sum_{i=1}^{k}\sqrt{\frac{s_{i} \log q + \log (1/\delta)}{N}},
\end{split}
\end{equation}
\end{footnotesize}
where $s_{i}$ denotes the number of parameters in each optimizing group $i$. 
\end{theorem}

\textit{Proof.} 
We first derive HiFT generalization bound between the objective with parameters after a first step of optimization at one training step $\mathcal{L}_{test}(\theta_{hift}^{(1)})$ and the objective with parameters that has the best test performance $\mathcal{L}_{test}(\theta^{*})$:
\begin{equation}
\small
\begin{split}
\mathcal{L}_{test} & (\theta_{hift}^{(1)})- \mathcal{L}_{test}(\theta^{*}) \\
& \leq 4 \epsilon \\
& + 2 \sup_{\Tilde{\theta}^{(1)}} | \mathcal{L}_{test}(\bar{q}(\Tilde{\theta}^{(1)})) - \mathcal{L}_{train}(\bar{q}(\Tilde{\theta}^{(1)})) | \\
& + \mathcal{L}_{test}(\theta^{(1) *}) - \mathcal{L}_{test}(\theta^{*}), 
\end{split}
\end{equation}
with probability at least $1- \delta$, the second term can be bounded:
\begin{footnotesize}
\begin{equation}
\begin{split}
2 \sup_{\Tilde{\theta}^{(1)}} | \mathcal{L}_{test}(\bar{q}(\Tilde{\theta}^{(1)})) - \mathcal{L}_{train}(\bar{q}(\Tilde{\theta}^{(1)})) |
\\ \leq 2\sqrt{\frac{s_{1} \log q + \log (1/\delta)}{N}}
\end{split}
\end{equation}
\end{footnotesize}
The above inequality can be shown by considering Theorem D.2 in~\citet{PanigrahiSZA23} and taking $\Theta_{N} = 1$. 

Similarly, we can have: 
\begin{footnotesize}
\begin{equation}
\begin{split}
    \mathcal{L}_{test} & (\theta_{hift}^{(i)}) - \mathcal{L}_{test}(\theta_{hift}^{(i-1)}) \\
    & \leq 4 \epsilon \\
    & + 2 \sup_{\Tilde{\theta}^{(i)}} | \mathcal{L}_{test}(\bar{q}(\Tilde{\theta}^{(i)})) - \mathcal{L}_{train}(\bar{q}(\Tilde{\theta}^{(i)})) | \\
    & + \mathcal{L}_{test}(\theta^{(i) *}) - \mathcal{L}_{test}(\theta_{hift}^{(i-1)})
\end{split}
\end{equation}
\end{footnotesize}
Summing over the above terms with $i = \{ 1, ..., k\}$ completes the proof of this theorem.

\section{Memory Analysis}
\label{sec:memory}
According to previous work~\citep{LOMO,abs230517333}, the main components that consume GPU memory during the fine-tuning process include the weight parameter, optimizer states, gradients, and calculation of residual states (i.e, activations, temporary buffers and fragmented memory)~\cite{rajbhandari2020zero}. 
In this section, we give theoretical analysis on the GPU memory advantages of HiFT strategy from the perspectives of \textbf{weight parameter}, \textbf{optimizer states} and \textbf{gradients}~\footnote{Since the GPU memory occupied by forward activations is related to the model implementation, batch size and sentence length, we analyze the GPU memory requirements of internal variables through experiments.}. 
Assuming the model is fine-tuned using the AdamW optimizer with 32-bit precision, we employ $\zeta_1$, $\zeta_2$, $\zeta_3$ to represent the GPU memory used by weight parameter, optimizer states and gradients respectively. AdamW optimizer stores the gradient first moment estimation and second moment estimation, which means that the optimizer state parameter 
$\zeta_2$ is two times larger than weight parameter $\zeta_1$ (i.e., $\zeta_2 = 2* \zeta_1$). The gradient parameters typically correspond to the parameters updated in the model (i.e., $\zeta_3 = \zeta_1$). Therefore, the number of gradient parameters $\zeta_3$ is the same as the number of parameters $\zeta_1$ that need to be updated in the model. Therefore, for standard FPFT, the GPU memory required for these parameters is as follows:
\begin{equation}
\begin{split}
\mathcal{\zeta}_{fpft} &= \zeta_1 + \zeta_2 + \zeta_3 \\
            & = \zeta_1 + 2 \zeta_1 + \zeta_1 \\
            & = 4 \zeta_1
\end{split}
\end{equation}
Taking the fine-tuning of a 7B model at 32 precision using the AdamW optimizer as an example, the $\zeta_1$ is about 26.08G. Theoretically, the GPU memory required for fine-tuning these three parts of the 7B model is approximately 104.32 GB. If considering GPU memory occupied by forward activations and the impact of batch size and sentence length, the actual scenario FPFT requires more GPU memory than 104.32 GB. Under the HiFT training strategy, since only one group of parameters is updated for each training step, only the gradients of the updated parameters and the corresponding optimizer states are stored in the GPU according to Algorithm 1. The weight parameter need to reside in the GPU memory for forward propagation. Therefore, the average GPU memory required for each training step is as follows:
\begin{equation}
\begin{split}
\mathcal{\zeta}_{hift} &= \zeta_1 + \frac{\zeta_2}{k} +\frac{\zeta_3}{k} \\
            &= \frac{k+3}{k} * \zeta_1
\end{split}
\end{equation}
Compared with FPFT, the memory saved by HiFT in model parameters, gradients and optimizer states is:
\begin{equation}
\begin{split}
\Delta \mathcal{\zeta} &= \mathcal{\zeta}_{fpft} - \mathcal{\zeta}_{hift} \\
                  &= \frac{3k-3}{k} * \zeta_1
\end{split}
\end{equation}
In addition to these computable fixed parameters, HiFT can reduce the number of activation-related parameters that simultaneously reside in memory, which is discussed in the experimental section.
Considering the embedding layer, task-related head layer and 32 hidden layers, LLaMA-7B has $n=34$ layers. When $m=1$, it can be deduced that $k=34$, and the required GPU memory can be inferred to be $\mathcal{\zeta}_{hift} \approx 31.13G$, the GPU memory saving is about 73.19G compared with FPFT.

\section{Baselines}
\label{sec:base}
\textbf{Language Models} include \textbf{RoBERTa}~\cite{abs190711692} with \textbf{base} and \textbf{large} versions, \textbf{GPT-2}~\cite{radford2019language} with \textbf{medium} and \textbf{large} versions, \textbf{LLaMA}~\cite{Llama2} with \textbf{7B} and \textbf{13B} versions, and \textbf{OPT-13B}~\cite{abs220501068}.

\textbf{Fine-Tuning strategies} include \textbf{BitFit}~\cite{zaken2022bitfit}, \textbf{Adapter}~\cite{houlsby2019parameter}, \textbf{Prefix}~\cite{lester2021power}, \textbf{LoRA}~\cite{hu2021lora}, \textbf{MeZO}~\cite{abs230517333}, \textbf{S4}~\cite{chen2023parameter}, \textbf{Adapter$^\text{L}$}~\cite{lin2020exploring}, \textbf{PreLayer}~\cite{hu2021lora}, IA3~\cite{liu2022few}, and \textbf{FPFT}.
\textbf{Optimizers} include \textbf{AdamW}~\cite{Loshchilov2017DecoupledWD}, \textbf{SGDM}~\cite{Qian99}, \textbf{SGD}, \textbf{Adafactor}~\cite{ShazeerS18}, \textbf{Adagrad}~\cite{DuchiHS10}. 
Some baselines might only appear in certain experiments.

\section{Datasets}
\label{sec:data}
We conduct experiments on the following datasets: 
\textbf{GLUE}~\cite{Wang2018GLUEAM} (SST-2~\cite{socher2013recursive}, 
    CoLA~\cite{warstadt2019neural}, MNLI~\cite{williams2018broad}, MRPC~\cite{warstadt2019neural}, QNLI~\cite{rajpurkar2018know}, QQP\footnote{https://quoradata.quora.com/First-Quora-Dataset-Release-Question-Pairs}, RTE and STS-B~\cite{cer2017semeval}); 
    \textbf{SuperGLUE} (CB~\cite{de2019commitmentbank}, BoolQ~\cite{clark2019boolq}, COPA~\cite{roemmele2011choice}, MultiRC~\cite{khashabi2018looking}, RTE, WiC~\cite{pilehvar2019wic}, WSC~\cite{levesque2012winograd}, ReCoRD~\cite{zhang2018record}), \textbf{SQuAD}~\cite{rajpurkar2016squad}, \textbf{E2E}~\cite{novikova2017e2e}, \textbf{DROP}~\cite{dua2019drop}, \textbf{ViGGO}~\cite{juraska2019viggo}, \textbf{SQL Generation}~\cite{yu2018spider,zhongSeq2SQL2017} and \textbf{GSM8K}~\cite{cobbe2021gsm8k}.

\section{Difference from Splitting Optimization}
The purpose of splitting optimization is to serve parallel computing. For example, matrix $C = A\cdot B$, matrix $A$ can be divided into A$_1$ and A$_2$ by row, then $C = [A_1 \cdot B; A_2 \cdot B]$. We can put $A_1 \cdot B $ and $A_2 \cdot B$ on different devices and calculate them in parallel. 
The purpose of HiFT is full-parameter model fine-tuning on low-resource devices. HiFT only updates a subset of parameters at each training step. Reduce the number of trainable parameters in each step through layer-by-layer asynchronous updates, thereby reducing the memory usage of fine-tuning models. Both the algorithm process and the purpose of the algorithm are different.

Besides, the theory behind splitting optimization is the matrix block principle. This principle states that a large matrix can be divided into smaller submatrices or blocks. These blocks can then be manipulated independently. The result of each block is a subset of the original matrix multiplication result. Megatron-LM applies the splitting optimization principle to conduct large-scale parallel training of language models. However, HiFT does not rely on the matrix block principle. HiFT's updates are independent at each step, not a subset of standard fine-tuning, and is a new approach independent of standard fine-tuning. The relationship between HiFT's update process and standard fine-tuning cannot be described using splitting optimization.

\section{Implementation Details}
\label{sec:setting}
The performance results of the experiment are based on training with the AdamW optimizer. For RoBERTa$_\text{base}$ and RoBERTa$_\text{large}$ models, we follow~\citet{chen2023parameter} for the hyperparameter setting of no-prompt fine-tuning such as batch size and learning rate. For GPT-2$_\text{medium}$and GPT-2$_\text{large}$, 
we follow~\citet{hu2021lora} for the hyperparameter setting for no-prompt fine-tuning such as batch size and learning rate. For RoBERTa$_\text{large}$ model, we follow~\citet{abs230517333} for the hyperparameter setting of prompt fine-tuning such as prompt template, batch size and learning rate. The specific model layering principle is that all embedding layers are treated as a single layer including position coding, all head layer parameters are treated as a single layer, and the remaining layers are divided according to the construction structure of the model. For example, RoBERTa$_\text{base}$ has 12 hidden layers, thus are divided into 12 layer units. Then we group them according to the divided layers. Table~\ref{tab:hift_hyper} reports hyperparameter used for HiFT. For instruction fine-tuning, we fine-tune these languages models on Alpaca dataset~\cite{taori2023stanford}. Alpaca contains 51K instruction-following demonstrations generated from text-davinci-003 (GPT-3.5). For evaluation, we use the fine-tuned models to generate responses for the pre-defined questions, which are from the MT-bench~\cite{zheng2024judging}. GPT-4 takes these answers as input and evaluates them with scores within 10. Repository FastChat\footnote{https://github.com/lm-sys/FastChat} provides the detailed evaluation process.

\begin{table*}[h]
\centering
\small
\begin{tabular}{lrc}
\toprule
Experiment & Hyperparameters & Values \\
\midrule
RoBERTa-base & Total Batch size & $64$ \\
& Learning rate & $\{1\mathrm{e}{-5}, 2\mathrm{e}{-5}, 3\mathrm{e}{-5} \}$ \\
& warmup & \{0.0, 0.02, 0.06\} \\
& Device & 8*GTX 1080Ti (11G) \\
& Weight Decay & $0$ \\
\midrule
RoBERTa-large & Total Batch size & $32$ \\
& Learning rate & $\{1\mathrm{e}{-5}, 2\mathrm{e}{-5}, 3\mathrm{e}{-5} \}$ \\
& warmup & \{0.0, 0.02, 0.06\} \\
& Device & 8*GTX 1080Ti (11G) \\
& Weight Decay & $0$ \\
\midrule
GPT-2 (M) & Batch size & $32$ \\
& Learning rate & $\{5\mathrm{e}{-5}\}$ \\
& warmup & \{0.0\} \\
& Device & RTX A6000 (48G) \\
& Temperature & 0.75 \\
& Beam size & 16 \\
& repetition penalty & 4 \\
& length penalty & 0.9 \\
\midrule
GPT-2 (L) & Batch size & $32$ \\
& Learning rate & $\{5\mathrm{e}{-5}\}$ \\
& warmup & \{0.0\} \\
& Device & RTX A6000 (48G) \\
& Temperature & 0.75 \\
& Beam size & 16 \\
& repetition penalty & 4 \\
& length penalty & 0.9 \\
\midrule
RoBERTa-large & Batch size ($k=16$) & $\{2,4,8\}$ \\
 & Batch size ($k=512$) & $\{8,16,32\}$ \\
& Learning Rates &  $\{1\mathrm{e}{-5}, 3\mathrm{e}{-5}, 5\mathrm{e}{-5}, 8\mathrm{e}{-5} \}$ \\
& Device & 8*GTX 1080Ti (11G) \\
& Weight Decay & $0$\\
\midrule
OPT-13B & Batch size & $\{2,4,8\}$ \\
& Learning Rates & $\{1\mathrm{e}{-5}, 2\mathrm{e}{-5}, 5\mathrm{e}{-5},8\mathrm{e}{-5}\}$ \\
& Device & A100 (80G) \\
& Weight Decay & $0$\\
\midrule
Mistral-7B & Batch size & $\{2,4,8\}$ \\
& Learning Rates & $\{1\mathrm{e}{-5}, 2\mathrm{e}{-5}, 5\mathrm{e}{-5}\}$ \\
& Device & A100 (80G) \\
& Weight Decay & $0$\\
\midrule
TinyLLaMA & Batch size & $\{2,4,8\}$ \\
& Learning Rates & $\{2\mathrm{e}{-5}, 5\mathrm{e}{-5},8\mathrm{e}{-5}\}$ \\
& Device & A100 (80G) \\
& Weight Decay & $0$\\
\midrule
LLaMA2-7B & Batch size & $\{2,4,8\}$ \\
& Learning Rates & $\{1\mathrm{e}{-5}, 2\mathrm{e}{-5}, 5\mathrm{e}{-5},8\mathrm{e}{-5}\}$ \\
& Device & A100 (80G) \\
& Weight Decay & $0$\\
\midrule
LLaMA2-13B & Batch size & $\{2,4,8\}$ \\
& Learning Rates & $\{1\mathrm{e}{-5}, 2\mathrm{e}{-5}, 5\mathrm{e}{-5},8\mathrm{e}{-5}\}$ \\
& Device & A100 (80G) \\
& Weight Decay & $0$\\
\bottomrule
\end{tabular}
\caption{The hyperparameter grids used for HiFT experiments.}
\label{tab:hift_hyper}
\end{table*}

\section{More Experiment Results}
\label{sec:results}

\begin{table*}[ht]
\begin{adjustbox}{max width=0.85\textwidth, center}
\includegraphics[width=\textwidth]{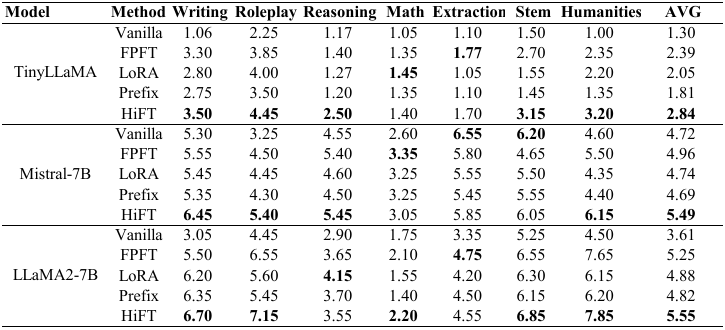}
\end{adjustbox}
\caption{Performance comparison of different fine-tuning methods on the MT-Bench. The rank of LoRA is 64, and the number of virtual words of prefix is 128.}
\label{tab:r-mt}
\end{table*}

\begin{figure*}[ht]
\centering
\includegraphics[width=0.9\textwidth]{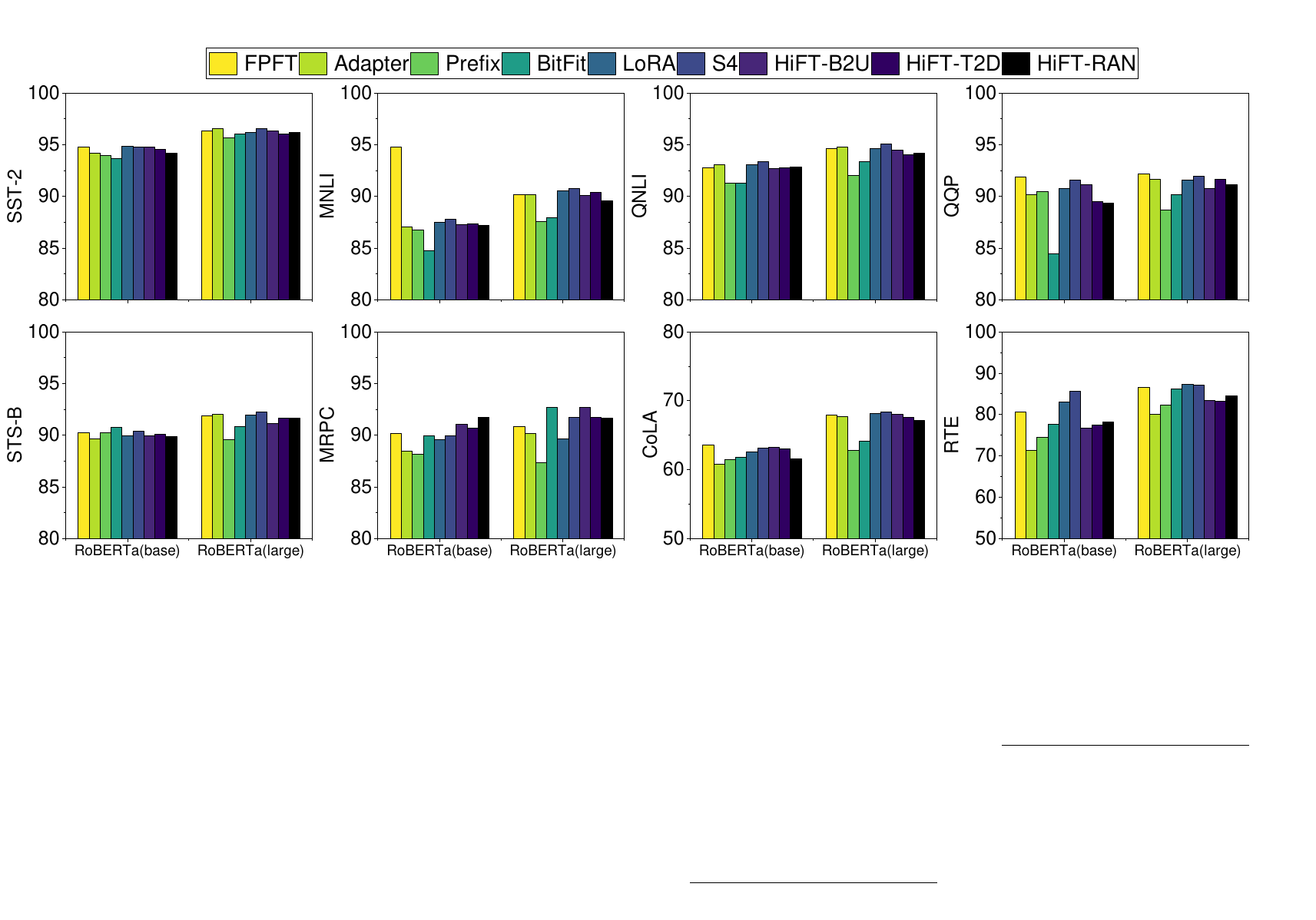}
\caption{RoBERTa results on different fine-tuning strategies. We report accuracy metrics for the SST-2, QNLI, QQP, MRPC and RTE, mean accuracy for MNLI, spearman coefficient for STS-B and matthews correlation coefficient for CoLA. The $m$ of HiFT is set to 1. 
B2U, T2D and RAN are bottom2up, top2down and random strategies.
}
\label{fig:roberta_result}
\end{figure*}

\subsection{Proportion of Parameters}
\begin{figure*}[!t]
\centering
\includegraphics[width=\textwidth]{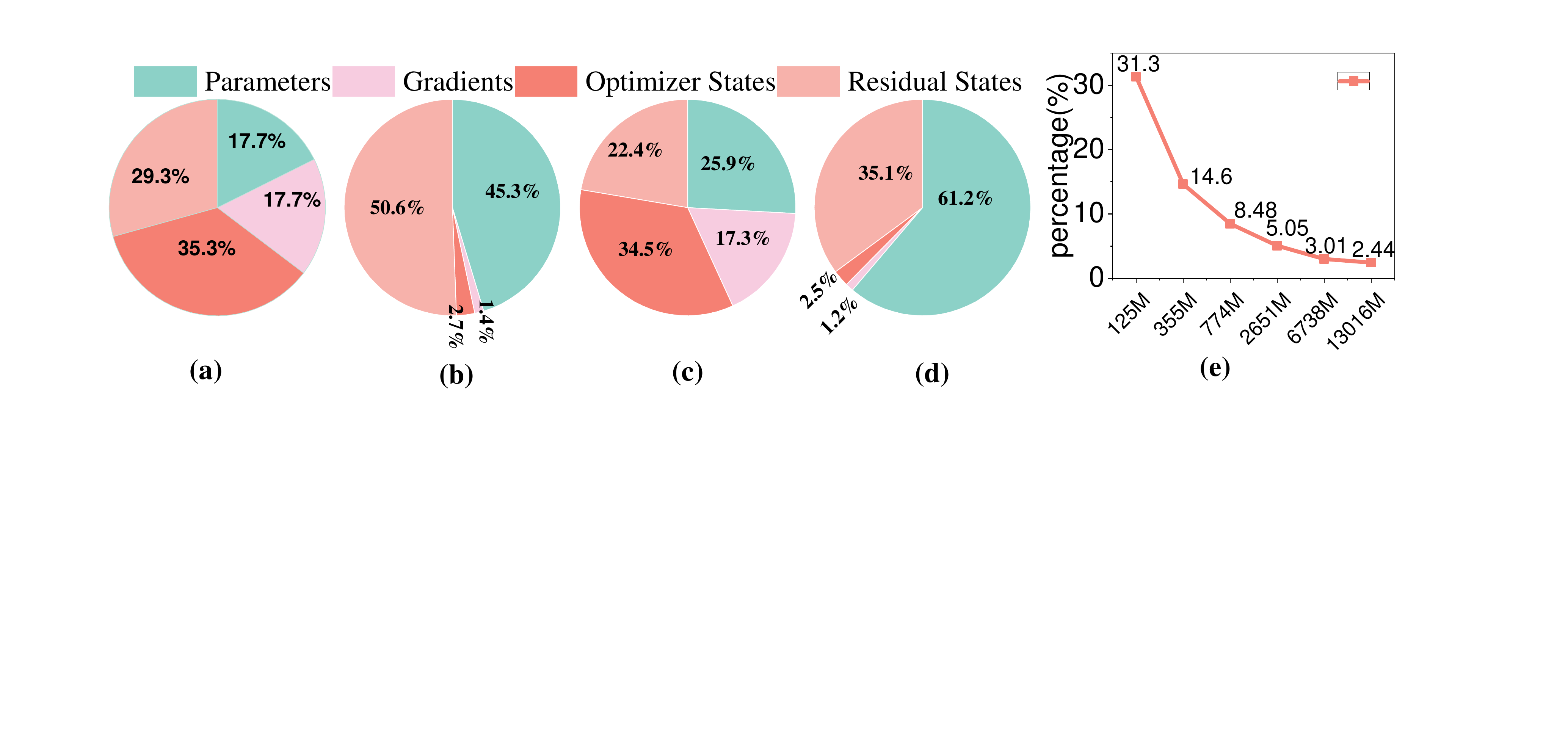}
\caption{(a), (b), (c) and (d) represent the proportion of parameters occupied by different parts when fine-tuning LLaMA-2 (7B). The sequence length and batch size are set to 512 and 6. (a): 32-bit precision FPFT; (b): 32-bit precision HiFT; (c) mixed precision FPFT; (d): mixed precision HiFT. Fine-tuning uses the AdamW optimizer. The $m$ is set to 1 for HiFT. (e) represents the change in the proportion of the peak trainable parameters to the total model parameters during the HiFT training under different size models.}
\label{fig:percentage}
\end{figure*}

Figure~\ref{fig:percentage} (a, b, c, d) shows the percentage of memory used by the parameters of each part when fine-tuning LLaMA-2 (7B) under standard FPFT and HiFT with the AdamW optimizer. Figure~\ref{fig:percentage} (e) repotrs the changes in the amount of peak fine-tuning parameters under HiFT at different model sizes.

\subsection{Mixing Precision}
We observe an interesting phenomenon when fine-tuning the GPT-Neo (2.7B) (Table~\ref{tab:m-neo} in Appendix~\ref{sec:results}) and LLaMA-2 (7B) (Table~\ref{tab:m-llama}) using mixed precision, the memory usage is higher than FPFT. We find that when using mixed precision fine-tuning, both single-precision and half-precision parameters of the model exist simultaneously. Therefore, the model parameters use more memory in mixed precision than in standard FPFT. Mixed precision mainly focuses on reducing the memory usage of activation states (i.e., residual states). When the model's own parameter size is large, the memory increase of the model parameters may be greater than the memory reduction of mixed precision (when the batch size is not large enough). Therefore, it may appear that the memory usage of mixed precision is greater than standard FPFT. Due to the large number of parameters of LLMs (large language models), it is difficult to use larger batch sizes, so it is difficult to bring out the advantages of mixed precision in the context of large models. HiFT is an optional, more efficient solution that maintains single-precision full-parameter fine-tuning while greatly reducing memory usage.
We would like to emphasize that the current mixed precision does not support hierarchical operations, so it cannot take advantage of HiFT.

To fully exploit the advantages of HiFT, we have adapted mixed precision to HiFT. That is, each step only moves the single-precision weight corresponding to the parameter that needs to be updated to the GPU (Mixed precision makes a single-precision backup of the weights of the half-precision model.).
Table~\ref{tab:m-llama} reports the memory profiling for LLaMA2-7B using adapted mixed precision. When using the AdamW optimizer, the adapted mixed precision for HiFT saves approximately 76.65\% of GPU memory. When the batch size is 1, fine-tuning the LLaMA-7B model on the E2E data set requires approximately \textbf{16.87G} of GPU memory, and fine-tuning the LLaMA-13B model requires approximately 31G of memory. This means that HiFT supports FPFT of a 7B model on a device with 24G GPU memory.

\begin{table*}[!t]
\begin{adjustbox}{max width=\textwidth, center}
\includegraphics[width=\textwidth]{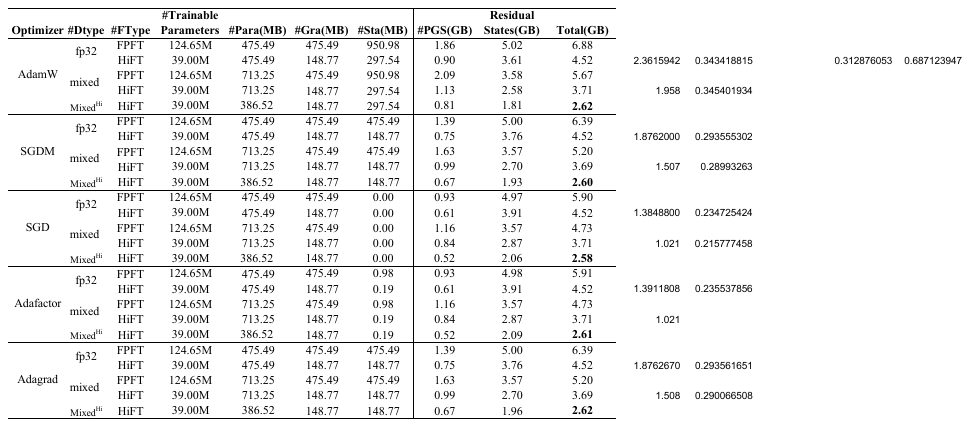}
\end{adjustbox}
\caption{The GPU memory usage of fine-tuning RoBERTa$_{\text{base}}$ on the CoLA dataset. The sequence length and batch size are set to 512 and 8, respectively. \textbf{\#Dtype} represents the data type used for training, where \textbf{FP32} represents fully parameter fine-tuning the model with 32-bit precision, and \textbf{mixed} represents fine-tuning with mixed precision. \textbf{\#Trainable parameters} represents the maximum number of trainable parameters that appear in a single step during the fine-tuning process. \textbf{\#Para} represents the memory occupied by the model parameters, \textbf{\#Gra} represents the memory occupied by the gradient, and \textbf{\#Sta} represents the memory occupied by the optimizer state. \textbf{\#PGS} represents the sum of memory occupied by model parameters (i.e.,\textbf{\#Para}), gradients (i.e.,\textbf{\#Gra}) and optimizer state (i.e.,\textbf{\#Sta}). \textbf{Residual states} mainly includes activation, temporary buffers and unusable fragmented memory. \textbf{Total} represents the total memory used during fine-tuning. The parameter $m$ of HiFT is set to 1.}
\label{tab:m-base}
\end{table*}
\begin{table*}[ht]
\begin{adjustbox}{max width=\textwidth, center}
\includegraphics[width=\textwidth]{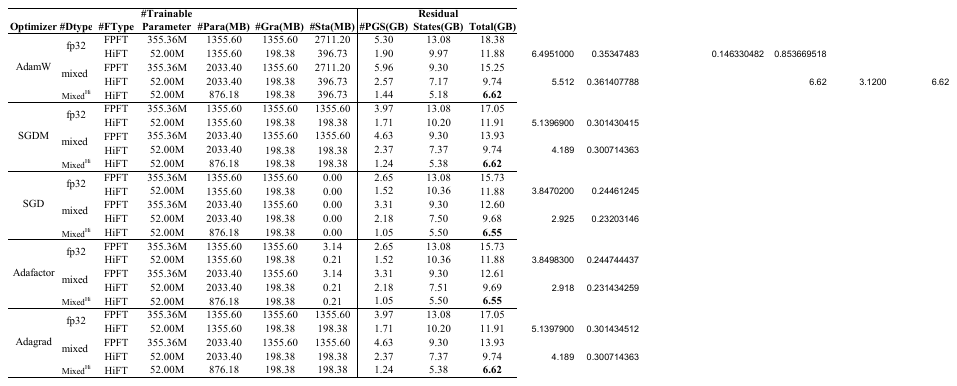}
\end{adjustbox}
\caption{The GPU memory usage of fine-tuning RoBERTa$_{\text{large}}$ on the CoLA dataset. The sequence length and batch size are set to 512 and 8, respectively.}
\label{tab:m-large}
\end{table*}
\begin{table*}[ht]
\begin{adjustbox}{max width=\textwidth, center}
\includegraphics[width=\textwidth]{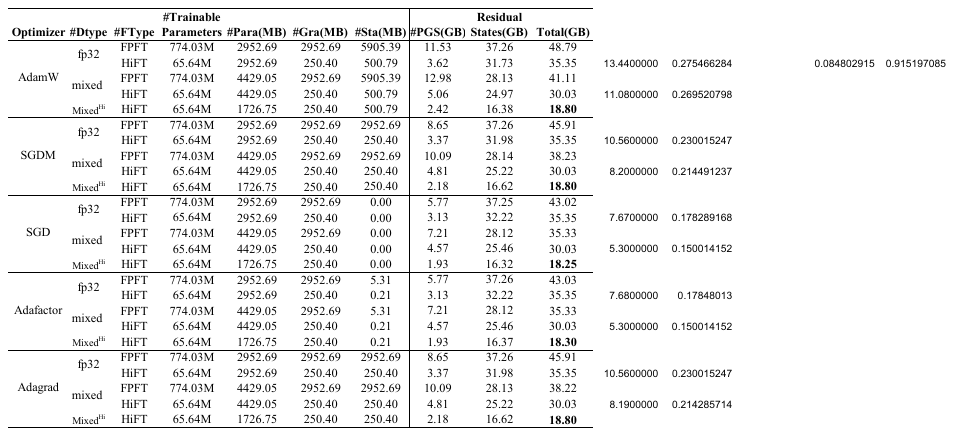}
\end{adjustbox}
\caption{The GPU memory usage of fine-tuning GPT-2$_{\text{large}}$ on the E2E dataset. The sequence length and batch size are set to 512 and 8, respectively.}
\label{tab:m-gptl}
\end{table*}

\begin{table*}[ht]
\begin{adjustbox}{max width=\textwidth, center}
\includegraphics[width=\textwidth]{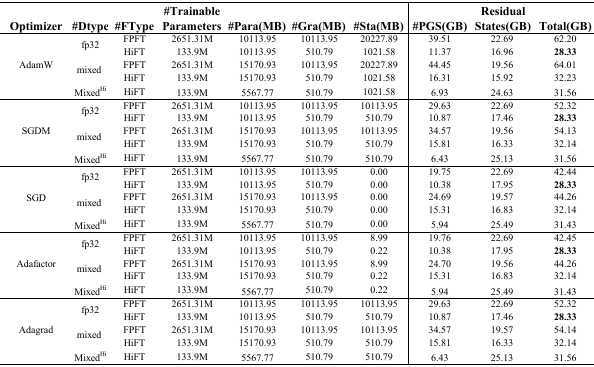}
\end{adjustbox}
\caption{The GPU memory usage of fine-tuning GPT-Neo on the E2E dataset. The sequence length and batch size are set to 512 and 8, respectively.}
\label{tab:m-neo}
\end{table*}

\begin{table*}[ht]
\begin{adjustbox}{max width=0.95\textwidth, center}
\includegraphics[width=\textwidth]{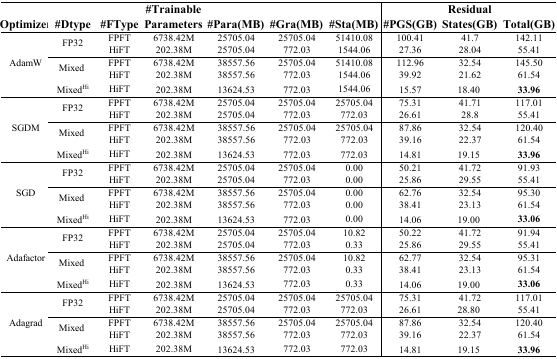}
\end{adjustbox}
\caption{The GPU memory usage of fine-tuning LLaMA (7B) on the E2E dataset. The sequence length and batch size are set to 512 and 6, respectively.}
\label{tab:m-llama}
\end{table*}



\subsection{Prompts}
Tables~\ref{tab:prompt1} and~\ref{tab:prompt_opt} gives detailed prompts of different datasets.

\label{sec:prompt}

\newcommand{\cls}{\ttt{[CLS]}}
\newcommand{\tableindent}{~~}
\newcommand{\mask}{\texttt{[MASK]}}
\newcommand{\firstsent}{\ttt{<}$S_1$\ttt{>}}
\newcommand{\secondsent}{\ttt{<}$S_2$\ttt{>}}
\newcommand{\sent}{\ttt{<}$S_1$\ttt{>}}
\newcommand\sys[1]{\textsc{#1}}
\newcommand\ti[1]{\textit{#1}}
\newcommand\ts[1]{\textsc{#1}}
\newcommand\ttt[1]{\texttt{#1}}

\begin{table*}[h]
\centering
\small
\begin{tabular}{lllll}
\toprule
 \tf{Dataset} & $C$  & \tf{Type} & \tf{Prompt} & \tf{Label words} \\
 \midrule
 SST-2 & 2 & sentiment  cls.& {\sent} It was {\mask} . & \{great, terrible\} \\
 SST-5 & 5  & sentiment cls.& {\sent} It was {\mask} . & \{great, good, okay, bad, terrible\} \\
 TREC & 6   & topic cls. & {\mask} : {\sent} & \{Description, Expression, Entity, \\
 & & & & Human, Location, Number\}\\
  MNLI & 3  &  NLI & {\firstsent} ? {\mask} , {\secondsent}  & \{Yes, Maybe, No\}\\
SNLI & 3   & NLI  & {\firstsent} ? {\mask} , {\secondsent} & \{Yes, Maybe, No\} \\
 RTE & 2  & NLI & {\firstsent} ? {\mask} , {\secondsent} & \{Yes, No\}  \\
\bottomrule
\end{tabular}
\caption{The  prompts of the datasets we used in our RoBERTa-large experiments (i.e., Table~\ref{tab:r-prompt}). 
The prompts are adapted from \cite{gao-etal-2021-making} and include a template and a set of label words that can fill in the \mask token. {\firstsent} and {\secondsent} refer to the first and the second (if any) input sentence. $C$ is the number of labels.}
\label{tab:prompt1}
\end{table*}

\newcommand{\nnn}{\textbackslash{}n}
\newcommand{\lword}[1]{{\color{blue}#1}}
\newcommand{\inp}[1]{\ttt{#1}}

\begin{table*}[h]
\centering
\small
\begin{tabular}{lll}
\toprule
 \tf{Dataset}  & \tf{Type} & \tf{Prompt}\\
 \midrule
 SST-2  &  cls.& {\inp{<text>}} It was \lword{terrible}/\lword{great} \\
 RTE & cls. & \inp{<premise>}\\
 &&Does this mean that "\inp{<hypothesis>}" is true? Yes or No?\\
 && \lword{Yes}/\lword{No} \\
CB & cls. & Suppose \inp{<premise>} Can we infer that "\inp{<hypothesis>}"? Yes, No, or Maybe? \\
&& \lword{Yes}/\lword{No}/\lword{Maybe}\\
BoolQ & cls. & \inp{<passage>} \inp{<question>}? \\
&& \lword{Yes}/\lword{No} \\
WSC & cls. & \inp{<text>}\\
&& In the previous sentence, does the pronoun "\inp{<span2>}" refer to \inp{<span1>}? Yes or No? \\
&& \lword{Yes}/\lword{No}\\
WIC & cls. & Does the word "\inp{<word>}" have the same meaning in these two sentences? Yes, No? \\
&& \inp{<sent1>}\\
&& \inp{<sent2>}\\
&& \lword{Yes}/\lword{No} \\
MultiRC & cls. & \inp{<paragraph>}\\
&& Question: \inp{<question>} \\
&& I found this answer "\inp{<answer}". Is that correct? Yes or No? \\
&& \lword{Yes}/\lword{No} \\
COPA & mch. &  \inp{<premise>} so/because \inp{<candidate>}\\
ReCoRD & mch. & \inp{<passage>}\\
& & \inp{<query>.replace("@placeholder", <candidate>)} \\
SQuAD & QA & Title: \inp{<title>}\\
&& Context: \inp{<context>}\\
&& Question: \inp{<question>}\\
&& Answer: \\
DROP & QA& Passage: \inp{<context>}\\
&& Question: \inp{<question>}\\
&& Answer: \\
\bottomrule
\end{tabular}
\caption{
The prompts of the datasets we used in our OPT experiments.  
There are three types of tasks: classification (cls.), multiple-choice (mch.), and question answering (QA). \inp{<text>} represents input from the dataset and \lword{Yes} represents label words. 
For inference on multiple choice tasks, we put in different candidates in the prompt and calculate the average log-likelihood for each candidate, and choose the candidate with the highest score. For inference on QA tasks, we use greedy decoding to generate the answer. All prompts configurations are consistent with ~\citet{abs230517333}
}
\label{tab:prompt_opt}
\end{table*}

\end{document}